\documentclass[10pt,twocolumn,letterpaper]{article}

\usepackage{cvpr}
\usepackage{times}
\usepackage{epsfig}
\usepackage{graphicx}
\usepackage{amsmath}
\usepackage{amssymb}
\usepackage{multirow}

\usepackage[breaklinks=true,bookmarks=false]{hyperref}

\cvprfinalcopy % *** Uncomment this line for the final submission

\def\cvprPaperID{****} % *** Enter the CVPR Paper ID here

\begin{document}

%%%%%%%%% TITLE
\title{Unsupervised Degradation Learning for Single Image Super-Resolution}

%% omit the following lines up until the closing ``}''.
%% Additional authors and addresses can be added with ``\and'',
%% just like the second author.
%% To save space, use either the email address or home page, not both
%\and
\author{Tianyu Zhao${^1}$, Wenqi Ren${^2}$, Changqing Zhang${^1}$, Dongwei Ren${^1}$, Qinghua Hu${^1}$\\
${^1}$School of Computer Science and Technology, Tianjin University\\
${^2}$Institute of Information Engineering, Chinese Academy of Sciences\\
}

\maketitle
%\thispagestyle{empty}

%%%%%%%%% ABSTRACT
\begin{abstract}
Deep Convolution Neural Networks (CNN) have achieved significant performance on single image super-resolution (SR) recently. However, existing CNN-based methods use artificially synthetic low-resolution (LR) and high-resolution (HR) image pairs to train networks, which cannot handle real-world cases since the degradation from HR to LR is much more complex than manually designed.
To solve this problem, we propose a real-world LR images guided bi-cycle network for single image super-resolution, 
in which the bidirectional structural consistency is exploited to train both the degradation and SR reconstruction networks in an unsupervised way.  
Specifically, we propose a degradation network to model the real-world degradation process from HR to LR via generative adversarial networks, and these generated realistic LR images paired with real-world HR images are exploited for training the SR reconstruction network, forming the first cycle. Then in the second reverse cycle, consistency of real-world LR images are exploited to further stabilize the training of SR reconstruction and degradation networks.  
%Both the degradation and the reconstruction networks are constrained for maintaining the structural consistency of the generated images and original images. 
%
Extensive experiments on both synthetic and real-world images demonstrate that the proposed algorithm performs favorably against state-of-the-art single image SR methods.
\end{abstract}

%%%%%%%%% BODY TEXT
\section{Introduction}

Single image super-resolution (SISR) aims to restore the high-resolution image from a single low-resolution image counterpart, which has been successfully used in many computer vision applications (\eg, medical imaging \cite{shi2013cardiac}, security monitoring \cite{zou2012very}, and image enhancement \cite{capel2000super}). Generally, a low-resolution image $y$ can be modeled as
\begin{equation} \label{eq:eps}
y = (x \otimes k)\downarrow _{\delta}+ g,
\end{equation}
where $x \otimes k$ is the convolution operation between the HR image $x$ and the blur kernel $k$, $\downarrow _{\delta}$ represents the operation of down-sampling image with scale factor of ${\delta}$, and $g$ denotes the Gaussian white noise. 
%${F(\cdot)}$ is the underlying mapping that reconstructs a HR image from a LR image.

%%%% figure 
\begin{figure}[t]
\begin{center}
\includegraphics[width=0.9 \linewidth]{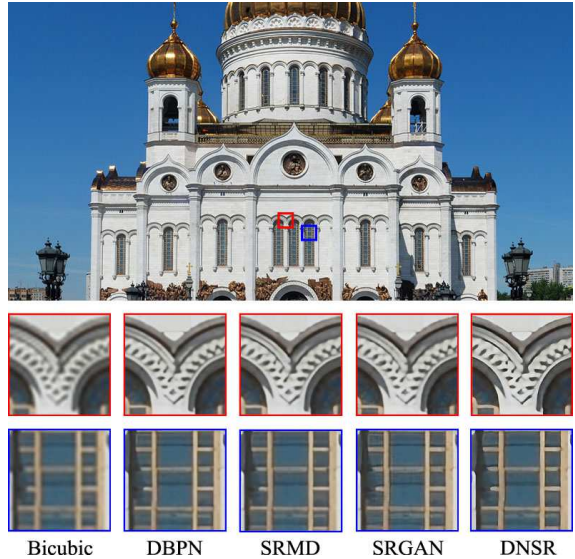}
\end{center}
\vspace{-3mm}
   \caption{${\times4}$ SR result for the real HR image `0879' (DIV2K). We directly reconstruct a higher-resolution image based on a real image. Our model (DNSR) can recover sharper edges and more details compared with other state-of-the-art methods.}
\label{fig 1}
\vspace{-5mm}
\end{figure}

Recently, a great number of methods have been proposed to learn the mapping between HR images and LR inputs \cite{dong2016image,dong2016accelerating,kim2016deeply,shi2016real,ledig2017photo,lim2017enhanced,zhang2018residual}. 
Dong \etal~\cite{dong2014learning} propose a CNN based image SR framework (SRCNN), which directly learns an end-to-end mapping to restore the HR image from a LR input by upsampling with bicubic interpolation first.
Kim \etal~\cite{Kim_2016_VDSR} design a pair of convolutional and nonlinear layers with gradient clipping to speed-up the training process, which outperforms SRCNN with a large margin thanks to stacked small filters and residual learning.
Lim \etal~\cite{lim2017enhanced} present an enhanced residual-block based network (EDSR) without normalization layer, which introduces a multi-scale architecture (MDSR) to handle multiple scales for various SR tasks. 
%Zhang \etal~\cite{zhang2018residual} propose a residual dense network (RDN) to make full use of each local layers and validates the advantage of learning complex features from LR images via a dense framework. 
%Further more, RDN extracts features from the LR images by proposing global feature fusion.

All the afore-mentioned CNN models are trained using synthesized LR images by the matched HR images.
However, it is difficult to obtain a realistic LR image by directly down-sampling a HR image.
% satisfactory result for these methods due to the pattern difference between the real LR image and the synthesized LR images. 
To model LR images in real cases, Zhang \etal~\cite{Zhang_2018_CVPR} propose a multiple degradations super-resolution network (SRMD) by taking the degradation maps and LR images as input to jointly consider noise, blur kernel, and down-sampler. 
Nevertheless, the noise level and blur kernel size are manually predefined, which weakens the ability to handle more general degradations and diverse LR images in real-world.
%

%WQ: put the following to realted work
%To diversify the texture of generated images, recently proposed methods of SRGAN \cite{ledig2017photo} and ESRGAN \cite{wang2018esrgan} introduce perceptual loss and GAN loss into the reconstruction network. Furthermore, CinCGAN  \cite{yuan2018unsupervised} resorts to unsupervised learning with unpaired data. But the LR images for training are generated artificially.
%
Since the patterns of real-world LR images and artificially degraded images have different characteristics, the models trained by synthesized LR images may be unpromising when applied to real-world LR images with complex combination of noise and blur, or for the case that LR image are obtained using different down-sampling methods.

To address the above limitations, inspired by the success of generative adversarial networks \cite{goodfellow2014generative} in image style translating \cite{CycleGAN2017}, we propose a novel unsupervised cycle super-resolution framework equipped with a degradation model to generate the realistic pattern in LR images, which acts as input of the reconstruction network. 
In this way, our model is applicable for complex degradation patterns rather than simple interpolations (\eg, bicubic and nearest-neighbor). As shown in Fig.~\ref{fig 1}, we directly reconstruct a higher-resolution image based on a real HR image with scale factor of 4. Our model can recover sharper edges and more details compared with other state-of-the-art methods (DBPN \cite{haris2018deep}, SRMD \cite{Zhang_2018_CVPR}, SRGAN \cite{ledig2017photo}).

This paper makes the following contributions:

\begin{itemize}

\item We propose an unsupervised learning network which consists of a degradation module and a reconstruction net. The degradation module is learned to generate realistic LR images for the reconstruction net in an unsupervised way. 
%
%a real LR image guided unsupervised bi-cycle degradation network for real image super-resolution, in which the degradation network can generate more realistic LR images for the reconstruction model.
%
%\item We propose a degradation-reconstruction cycle training strategy, in which the input to train our model are HR images unpaired with real LR images.
%
\item The process of generating LR images does not rely on the widely used down-sampling strategy. We introduce structure perceptual loss in the degradation network to preserve the structural similarity of generated LR images and the corresponding HR images. 
\item We develop a novel bi-cycle structure, where one cycle is designed for enforcing structural consistency between the degradation and SR reconstruction networks in an unsupervised way, and the other further stabilizes the training of SR reconstruction and degradation networks.
\item Extensive experiments on benchmark datasets and real-world images demonstrates that the proposed algorithm performs favorably against the state-of-the-art SR methods.

\end{itemize}
% end of introduction 
%-------------------------------------------------------------------------
\begin{figure*}
\begin{center}
%\fbox{\rule{0pt}{2in} \rule{.9\linewidth}{0pt}}
\includegraphics[width=0.9\linewidth]{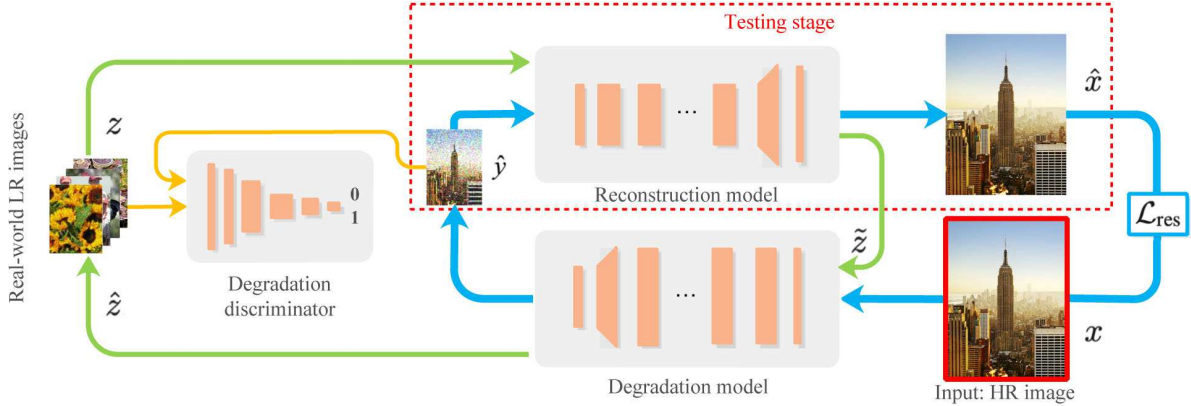}
\end{center}
\vspace{-3mm}
   \caption{Overview of the proposed DNSR network. For the cycle with blue arrows, given the input HR image $x$, ${\hat{y}}$ is the generated realistic LR image with the degradation model, based on which the HR image ${\hat{x}}$ is reconstructed with the reconstruction model. $\mathcal{L}_{\text{res}}$ is the loss for reconstruction model. For the cycle with green arrows, given a real-world LR image ${z}$, ${\tilde{z}}$ is the HR image generated by the reconstruction model and ${\hat{z}}$ is the generated realistic LR image degraded from ${\tilde{z}}$. The degradation discriminator enhances the probability that ${\hat{y}}$ is a real LR image. For testing, only reconstruction model is in used, with the input real LR images.}
\label{fig 2 framework}
\vspace{-5mm}
\end{figure*}
%% related works 
\section{Related Works}

In this section, we briefly review non-blind SISR and related blind SISR methods.

\subsection{Non-Blind SISR}

Early methods \cite{allebach1996edge,li2001new,zhang2006edge} super-resolve images based on the the interpolation-based theory. However, it is difficult to reconstruct detailed textures in the super-resolved results. 
%there are a lot of limitations when reconstructing realistic images with more details in texture. 
%
Dong \etal~\cite{dong2016image} propose a pioneer 3-layer CNN (SRCNN) for bicubic up-sampled image SR, which then brings outs a series of CNN-based SISR methods with more descent effectiveness and higher efficiency. 
On one hand, more effective CNN architectures are designed to improve SR performance, including very deep CNN with residual learning ~\cite{Kim_2016_VDSR} , residual and dense block \cite{ledig2017photo,lim2017enhanced}, recursive structure~\cite{kim2016deeply,Tai2017Image} and channel attention~\cite{zhang2018rcan}. 
On the other hand, separate research efforts are paid to speed up computational efficiency, where deep features are extracted from original LR image~\cite{dong2016accelerating, lai2017deep, shi2016real}. 
Taking both effectiveness and efficiency into account, this speed up strategy has also been succesively adopted in ~\cite{ledig2017photo,lim2017enhanced,Zhang_2018_CVPR,zhang2018rcan}.  

Recently, SRGAN \cite{ledig2017photo} and ESRGAN \cite{wang2018esrgan} introduce perceptual loss and adversarial loss into the reconstruction network. Spatial feature transform ~\cite{wang2018sftgan} are suggested to enhance texture details for photo-realistic SISR. Furthermore, CinCGAN \cite{yuan2018unsupervised} resorts to unsupervised learning with unpaired data. 
These methods, however, are all tailored to specific bicubic down-sampling, and usually perform limited on real-world LR images.
Although SRMD~\cite{Zhang_2018_CVPR} can handle multiple down-samplers by taking degradation parameters as input, these degradation parameters should be accurately provided, limiting its practical applications.  

In contrast, our proposed unsupervised degradation network could effectively model complex down-samplers and degradations learned from real-world LR training samples.

\subsection{Blind SISR}

Albeit there exist diverse degradations in real SISR applications, blurring is one of the vital aspect in degradation.
There are several successive work \cite{wang2005patch,michaeli2013nonparametric,shao2015simple} to estimate blur kernels from LR images, in which blurring and down-sampling are considered in the degradation model. 
But these methods rely on hand-crafted image priors and are also limited to diverse degradations. 
Recently, motivated by CycleGAN~\cite{CycleGAN2017}, several deep CNN-based methods are suggested to learn blind SR from unpaired HR-LR images. 
Yuan \etal~\cite{yuan2018unsupervised} present a Cycle-in-Cycle network to learn SISR and degradation models, but the degradation model is deterministic, making it limited in generating diverse and real-world LR images.

Closest to ours is the work of Bulat \etal \cite{bulat2018learn} in which the authors learn a high-to-low GAN to degrade and down-sample HR images, and then employ the LR-HR pairs to train a low-to-high GAN for blind SISR. Our method differs from \cite{bulat2018learn} in several important ways. 
%WQ: finished it. try to state three differences from [3].
First, both the structural consistency between the LR and HR images, and the relationship between reconstruction and degradation are explored by our bi-cycle structure, which jointly stabilizes the training of SR reconstruction and degradation networks. Second, since there are no pairs of LR-HR images in practice, our degradation model is trained in an unsupervised way, \ie, without using paired images.
We introduce unpaired real-world LR images into the GAN model for generating realistic LR images, and also exploit them to enhance the reconstruction model and degradation model jointly in a cycle.
%Bulat \etal~\cite{bulat2018learn} first utilize GAN to learn a High-to-Low GAN to degrade and downsample HR images, and then employ the LR-HR image pairs to train a Low-to-High GAN for blind SISR. 
%But the two GANs are separately trained, which is difficult to converge.   
%

In our bi-cycle degradation network, the bi-cycle consistency of LR images and HR images stabilize the training of both High-to-Low GAN and Low-to-High SR network, further boosting the superior SR performance.

%\subsection{Unsupervised learning}

% end of related works 
%-------------------------------------------------------------------------

%%% Proposed Methods
\section{Proposed Method}
% Unsupervised Degradation Learning for Single Image Super-Resolution
In this section, we present the unsupervised degradation learning for single image super-resolution, which effectively learns to generate LR images with realistic noise and blur patterns. We refer to this framework as Degradation Network for Super-Resolution (DNSR).
%Therefore, our model could be more reasonable for the task of real-world LR images SR and preserving details of images.

%% Structure overvie
\subsection{Overview of DNSR}
The proposed DNSR network architecture is illustrated in Fig.~\ref{fig 2 framework} which consists of the following three models: the degradation module, degradation discriminator, and reconstruction model. 
The degradation module aims to model the real-world degradation process from HR to LR images, and thus generates realistic LR images. 
The degradation discriminator is employed to ensure the degraded pattern in generated LR images to be similar to the real case. With the generated realistic LR and the corresponding HR images, the reconstruction model is trained to recover real structures and textures in HR images. 

Specifically, given a HR image ${x}$ as input, the degradation model down-samples it into a LR image $\hat{y}$, accordingly, the reconstruction model tries to recover the corresponding HR image $\hat{x}$ that is approximates ${x}$. This process is shown as the blue circle in Fig.~\ref{fig 2 framework}. 
To fully exploit the real-world LR images ${z}$, we used them in two ways. First, they are used to train the discriminator to promote the similarity between synthesized LR images and real ones. 
Second, as shown by the green circle, the real-world LR images are input into the reconstruction model to generate synthesized HR images which in turn act as input into the degradation model to reconstruct the original real-world LR images. This CycleGAN inspired manner further jointly enhances the relationship between  the reconstruction model and degradation model.
%Second, as shown by the green circle, the real LR images are input into the reconstruction model to generate synthesized HR images ${\tilde{z}}$ which acts as input into the degradation model to reconstruct the original real LR images ${z}$.

Different from previous work \cite{bulat2018learn,yuan2018unsupervised}, the LR image generated by our degradation model has no paired manually generated LR image as the ground-truth.

\begin{figure*}
\begin{center}
\includegraphics[width=0.9\linewidth]{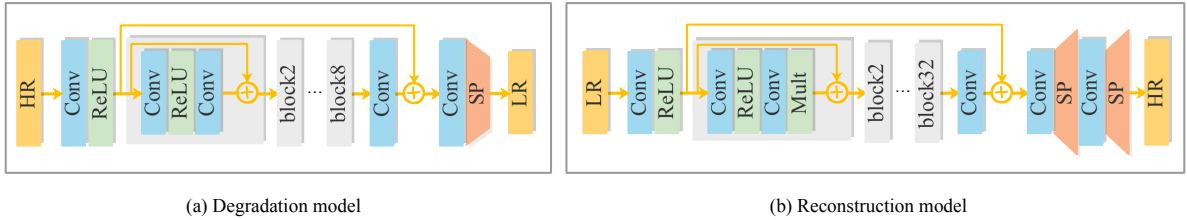}\\
\end{center}
\vspace{-3mm}
   \caption{The architecture of the proposed degradation model (a) and reconstruction model (b). Conv, ReLU and SP indicate the convolution layer composition, activation function, and sub-pixel convolution layer, respectively.}
\label{fig 3 architecture}
\vspace{-5mm}
\end{figure*}

%% Degradation model 

\subsection{Degradation Model}
%6
%WQ: you have stated the motivation in the Introduction, no need to restate it here. 
%The quality of input LR images will directly affect the performance of the super-resolution reconstructed model. However, it is difficult to obtain real LR-HR image pairs and generating synthesized LR images that are as close as real-world images are of vital importance. Moreover, the directly down-sampled LR images are not such close to real cases due to the manually designed way. In order 
%
To obtain more realistic LR images, we propose to model the mapping process from HR to real-world LR images by jointly using the degradation model and degradation discriminator.
The degradation discriminator aims to distinguish whether a LR image generated by the degradation model is close to real-world LR images. Degradation model in turn tries to generate more realistic images to fool the degradation discriminator. 
Different from the architecture proposed in SRGAN \cite{ledig2017photo}, our degradation discriminator enforces the generated LR image be similar to the real-world LR images instead of synthesized LR images.

%WQ: rubbish
%Deep CNNs have shown its power for image super-resolution \cite{dong2014learning,kim2016deeply}. In our work, we model the degradation process from a HR image to a LR image using a deep convolutional neural network, same as restoring a LR image to a HR image.

Figure~\ref{fig 3 architecture}(a) shows the architecture of the proposed degradation model.
Specifically, we employ one convolution layer with the ReLU \cite{krizhevsky2012imagenet} activation function as the first layer, and 8 residual blocks as the middle layers. We employ a convolution layer instead of the conventional down-sampling method, with the stride size ${s = 2}$ and ${s = 4}$ for scale factor of 2 and 4, respectively. We set kernel size as ${k = 3}$, number of filters as ${n = 32}$ for each convolution layer, and stride size as ${s = 1}$ for convolutional layers before the last one. 
We formulate the the degradation model as
\begin{equation} \label{Degradation model}
{\hat{y}} = G(x),
\end{equation}
where ${\hat{y}}$ is the LR image generated by the degradation model ${G(\cdot)}$ .

The degradation model outputs the LR image ${\hat{y}}$ which tries to fool the discriminator and thus induces the GAN loss as
\begin{equation} \label{GAN loss}
\mathcal{L}_{\text{adv}} = \frac{1}{N} {\sum_{i=1}^{N}} - \log D(G(x_i)),
\end{equation}
%
%WQ: exlain the symbol of "n". 
where ${D(\cdot)}$ represents degradation discriminator and $N$ is the number of input image patches.
%
%As recognized by previous work \cite{johnson2016perceptual}, the perceptual loss is rather effective in image style transformation. Degrading a HR image into a LR image can also be considered as one type of style transformation, \ie, from detailed images to coarse images. Moreover, 
In addition, since it is difficult to preserve the structure similarity between the generated LR and HR pair by using only the GAN loss, we introduce a structural perceptual loss \cite{johnson2016perceptual} to ensure the consistency in structure, and the loss is defined as
\begin{equation} \label{perceptual loss}
{\mathcal{L}_{\text{per}}} = \frac{1}{N} {\sum_{i=1}^{N}} {\parallel P_j(\hat{y_i})-P_j({x_i}) \parallel_2},
\end{equation}
where $\hat{y}$ is the generated realistic LR image by the degradation model, ${x}$ denotes the real HR image, ${P_j(\cdot)}$ is the ${j}^{th}$ maxpooling layer of the pre-trained VGG network \cite{simonyan2014very}. 
For matching the input size of VGG19 network, $\hat{y}$ and ${x}$ are scaled to the same size. 
%

%WQ: explain more about 4-th layer, edge? structure? low-level features?
ESRGAN \cite{wang2018esrgan} shows the difference of features obtained by  different layers of VGG19 network. The $4^{th}$ convolution layer before the $5^{th}$ maxpooling layer representing high-level features. The $2^{th}$ convolution layer before the $2^{th}$ maxpooling layer representing low-level features which contains more edges. The features of generated realistic LR image should be more blurry on the edges, close to real-world LR images. 
So, different from the conventional perceptual loss in ESRGAN , we use the $4^{th}$ convolution layer before the $4^{th}$ maxpooling layer as the output of ${P_j(\cdot)}$ since there is no need to obtain more details for perceiving texture features of LR images in our degradation task.
%
%The near-high-level features 

\begin{figure}[t]
\begin{center}
\includegraphics[width=0.9 \linewidth]{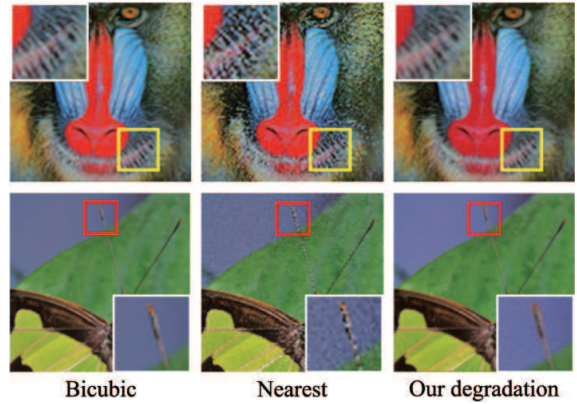}
\end{center}
\vspace{-3mm}
   \caption{Comparison on different degradations. The patterns of noise and blur are different between the LR images degraded with our degradation 	model and those of using bicubic and nearest-neighbor degradations. For all these methods, the down-sampling scale factor is 4.}
\label{fig 4 Compare on different degradation}
\label{fig:onecol}
\vspace{-5mm}
\end{figure}

%WQ: explain more, e.g., looks more blurry on the edges, close to real-world LR images.
Figure~\ref{fig 4 Compare on different degradation} shows the comparison for different types of degradation with down-sampling scale factor of 4. As shown, the generated LR images by our model contain different pattern of noise and blur compared with those of bicubic and nearest-neighbor degradations.

%%%% figure 

%% reconstruction model

\subsection{Reconstruction Model}
The structure of the proposed reconstruction model is demonstrated in Figure~\ref{fig 3 architecture}(b). 
%The similar structure as used in the degradation model with 32 residual blocks is employed. 
%WQ: I cannot understand this sentence. what is "with residual scaling factor 1 for training $\times2$". rewrite it. or you need to cite a reference for the conception of residual scaling factor.
We set ${k = 3}$, ${n = 64}$ and ${s = 1}$ for each convolution layer with residual scaling factor 1 for training $\times2$ SR model. For training $\times4$ SR model, we set ${n = 256}$ with residual scaling factor 0.1  \cite{lim2017enhanced}. 
Following \cite{shi2016real}, we use sub-pixel convolution layer for up-sampling to avoid the checkerboard artifacts \cite{odena2016deconvolution}. 
Note that we use the realistic LR images generated by our degradation model as inputs to ensure the reconstruction model can reconstruct HR images from real-world LR images. The reconstruction model is formulated as
\begin{equation} \label{Reconstruction model}
{\hat{x}} = R(\hat{y}),
\end{equation}
where ${\hat{x}}$ is the HR image generated by the proposed reconstruction model ${R(\cdot)}$.
To enforce the local smoothness and eliminate artifacts in restored images, we introduce a total variation loss as
\begin{equation} \label{TV loss}
{\mathcal{L}_{\text{tv}}} = \frac{1}{N} {\sum_{i=1}^{N}}( \parallel \nabla_h R(\hat{y_i})\parallel_2 + \parallel \nabla_v R(\hat{y_i})\parallel_2),
\end{equation}
where ${\nabla_h}$ and ${\nabla_v}$ are gradients of ${R(\hat{y_i})}$ in terms of horizontal and vertical directions, respectively.
In our model, we employ $L_1$ loss for the reconstruction loss and have the following formulation
\begin{equation} \label{L1 loss}
{\mathcal{L}_{1}} = \frac{1}{N} {\sum_{i=1}^{N}} \parallel R(\hat{y_i})-x_i \parallel_1,
\end{equation}
where ${x_i}$ is the ground-truth HR image.

%% degradation and reconstruction consistency 
\subsection{Degradation and Reconstruction Consistency}
To further jointly improve the reconstruction and degradation models,
%To further jointly improve the reconstruction model to restore HR images from real LR images with the same color space and the degradation model to generate more realistic LR images, 
we introduce a cycle consistency loss as shown with the green circle in Figure~\ref{fig 2 framework}. In this circle, a real-world LR image ${z}$ is taken as the input of our reconstruction model to generate a HR image ${\tilde{z}}$. Then, the degradation model tries to degrade the generated HR image ${\tilde{z}}$ to a realistic LR image ${\hat{z}}$. To ensure the generated realistic LR ${\hat{z}}$ is similar to the real-world LR image ${z}$, the cycle consistency loss is formulated as
\begin{equation} \label{cycle loss}
{\mathcal{L}_{\text{cyc}}} = \frac{1}{N} {\sum_{i=1}^{N}}{\parallel G(R(z_{i}))-z_{i} \parallel_1}.
\end{equation}

To jointly assure these effects mentioned above, the loss for the degradation model is induced as
\begin{equation} \label{deg loss}
{\mathcal{L}_{\text{deg}}} = {\mathcal{L}_{\text{cyc}}} + {\alpha \mathcal{L}_{\text{adv}}} + {\beta \mathcal{L}_{\text{per}}},
\end{equation}
where ${\alpha>0}$ and ${\beta>0}$ are tradeoff factors.

Considering the cycle consistency, the loss for the reconstruction model is induced as
\begin{equation} \label{rec loss}
{\mathcal{L}_{\text{rec}}} = {\mathcal{L}_{1}} + {\eta \mathcal{L}_{\text{cyc}}} + {\gamma \mathcal{L}_{\text{tv}}},
\end{equation}
where ${\eta>0}$ and ${\gamma>0}$ are tradeoff factors.
Accordingly, for the proposed DNSR model, we should optimize the following objective function
\begin{equation} \label{total loss}
{\mathcal{L}_{total}} = {\mathcal{L}_{deg}} + {\mathcal{L}_{rec}}.
\end{equation}

%%% Table 1 bicubic 
\begin{table*}[ht]
	\caption{SR results for bicubic degradation in terms of PSNR (dB) and SSIM. The values in red and blue indicate the best and second performances, respectively.}
	\vspace{-3mm}
	\small
	\begin{center}
		%\small
		\begin{tabular}{|c|c|c|c|c|c|c|c|c|}
			\hline
			
			Dateset& Scale& Bicubic & LapSRN & DBPN & SRMD & SRGAN & ESRGAN & DNSR \\
			
			\hline
			
			\multirow{2}*{Set5} & ${\times}$2 & 
			33.65 / 0.930&	
			37.52 / 0.959&	
			\color{red}38.09 \color{black}/ \color{blue}0.960&	
			37.53 / 0.959&	
			37.22 / 0.926&	
			37.81 / 0.953&	
			\color{blue}38.05 \color{black}/ \color{red}0.961 \\
			
			& ${\times}$4 & 
			28.42 / 0.810&	
			31.54 / 0.885&	
			\color{blue}31.75 \color{black}/ \color{red}0.898&
			31.59 / 0.887&	
			29.40 / 0.847&	
			31.40 / 0.871&	
			\color{red}31.76 \color{black}/ \color{blue}0.891 \\
			\hline
			\multirow{2}*{Set14} & ${\times}$2 & 
			30.34 / 0.870&	
			33.08 / 0.913&	
			\color{red}33.85 \color{black}/ \color{blue}0.919&	
			33.12 / 0.914&	
			32.14 / 0.886&	
			33.62 / 0.915&	
			\color{blue}33.83 \color{black}/ \color{red}0.922 \\
			
			& ${\times}$4 & 
			26.00 / 0.703&	
			28.19 / 0.772&	
			\color{red}28.34 \color{black}/ \color{blue}0.775&
			28.15 / 0.772&	
			26.64 / 0.710&	
			27.98 / 0.762&	
			\color{blue}28.33 \color{black}/ \color{red}0.776 \\
			\hline
			\multirow{2}*{Urban100} & ${\times}$2 & 
			26.88 / 0.841&	
			30.41 / 0.910& 
			\color{red}33.02 \color{black}/ \color{red}0.931&	
			31.33 / 0.920&	
			31.02 / 0.895&	
			32.01 / 0.913&	
			\color{blue}32.99 \color{black}/ \color{blue}0.928 \\
			
			& ${\times}$4 & 
			23.14 / 0.658&	
			25.21 / 0.756&	
			\color{blue}25.68 \color{black}/ \color{blue}0.785&	
			25.34 / 0.761&	
			25.11 / 0.725&	
			25.31 / 0.756&	
			\color{red}25.69 \color{black}/ \color{red}0.788 \\
			\hline
			\multirow{2}*{BSD100} & ${\times}$2 & 
			29.56 / 0.844&	
			31.80 / 0.895&	
			\color{red}32.27 \color{black}/ \color{blue}0.900&	
			32.05 / 0.898&	
			31.89 / 0.876&	
			31.99 / 0.887&	
			\color{blue}32.24 \color{black}/ \color{red}0.901 \\
			
			& ${\times}$4 & 
			25.96 / 0.668&	
			27.32 / 0.728&	
			\color{red}27.64 \color{black}/ \color{blue}0.740&	
			27.34 / 0.728&	
			25.16 / 0.668&	
			27.21 / 0.712&	
			\color{blue}27.61 \color{black}/ \color{red}0.742 \\
			\hline
			\multirow{2}*{DIV2K} & ${\times}$2 & 
			31.01 / 0.939&	
			34.35 / 0.942&	
			\color{blue}34.82 \color{black}/ \color{red}0.947&	
			34.73 / 0.940&	
			33.51 / 0.939&	
			33.69 / 0.941&	
			\color{red}34.83 \color{black}/ \color{blue}0.944 \\
			
			& ${\times}$4 & 
			26.66 / 0.852&	
			28.75 / 0.859&
			\color{red}28.94 \color{black}/ \color{red}0.869&	
			28.72 / 0.856&	
			28.09 / 0.821&	
			28.68 / 0.853&	
			\color{blue}28.87 \color{black}/ \color{blue}0.865 \\

			\hline
		\end{tabular}
	\end{center}
	\label{table: bicu}
	\vspace{-4mm}
\end{table*}

%%%  %%% Table 2 Nearest-neighbor  
\begin{table*}[ht]
	\caption{SR results for nearest-neighbor degradation in terms of PSNR (dB) and SSIM. The values in red and blue indicate the best and second performances, respectively.}
	\vspace{-3mm}
	\small
	\begin{center}
		%\small
		\begin{tabular}{|c|c|c|c|c|c|c|c|}
			\hline
			
			Dateset& Scale & LapSRN & DBPN & SRMD & SRGAN & ESRGAN & DNSR \\
			
			\hline
			
			\multirow{2}*{Set5} & ${\times}$2 & %29.06 / 0.880&
			\color{blue}26.23 \color{black}/ \color{blue}0.826&
			26.12 / 0.813&
			26.18 / 0.819&
			26.19 / 0.806&
			22.56 / 0.697&
			\color{red}26.25 \color{black}/ \color{red}0.828 \\
			
			& ${\times}$4 & %26.03 / 0.778&
			\color{blue}22.34 \color{black}/ \color{blue}0.716&
			22.15 / 0.680&
			22.28 / 0.712&
			21.79 / 0.713&
			21.53 / 0.479&
			\color{red}22.37 \color{black}/  \color{red}0.718 \\
			\hline
			\multirow{2}*{Set14} & ${\times}$2 & %27.70 / 0.838&
			\color{blue}25.19 \color{black}/ \color{blue}0.779&
			25.15 / 0.777&
			25.17 / 0.778&
			25.16 / 0.763&
			21.45 / 0.649&
			\color{red}25.21 \color{black}/  \color{red}0.782 \\
			
			& ${\times}$4 & %24.80 / 0.724&
			\color{blue}21.62 \color{black}/ \color{blue}0.657&
			21.56 / 0.651&
			21.57 / 0.654&
			21.02 / 0.587&
			17.12 / 0.361&
			\color{red}21.65 \color{black}/  \color{red}0.661 \\
			\hline
			\multirow{2}*{Urban100} & ${\times}$2 & %23.97 / 0.781&
			\color{blue}21.18 \color{black}/ \color{blue}0.715&
			20.99 / 0.703&
			21.12 / 0.712&
			20.94 / 0.698&
			17.47 / 0.583&
			\color{red}21.22 \color{black}/  \color{red}0.719 \\
			
			& ${\times}$4 & %19.92 / 0.538&
			\color{blue}16.97 \color{black}/ \color{blue}0.455&
			16.37 / 0.439&
			16.95 / 0.438&
			16.03 / 0.398&
			12.65 / 0.204&
			\color{red}17.01 \color{black}/  \color{red}0.457 \\
			\hline
			\multirow{2}*{BSD100} & ${\times}$2 & %27.03 / 0.793&
			\color{blue}24.13 \color{black}/ 0.725&
			24.02 / 0.718&
			24.11 / \color{blue}0.726&
			23.87 / 0.705&
			20.15 / 0.624&
			\color{red}24.19 \color{black}/  \color{red}0.732 \\
			
			& ${\times}$4 & %22.15 / 0.567&
			\color{blue}19.01 \color{black}/ \color{blue}0.483&
			18.53 / 0.467&
			18.85 / 0.474&
			18.29 / 0.421&
			13.90 / 0.183&
			\color{red}19.06 \color{black}/  \color{red}0.486 \\
			\hline
			\multirow{2}*{DIV2K} & ${\times}$2 & %29.52 / 0.865&
			26.88 / 0.814&
			26.16 / 0.798&
			\color{blue}26.89 \color{black}/ \color{blue}0.818&
			26.25 / 0.789&
			21.56 / 0.661&
			\color{red}26.91 \color{black}/  \color{red}0.826 \\
			
			& ${\times}$4 & %24.93 / 0.653&
			22.13 / 0.579&
			21.65 / 0.569&
			\color{blue}22.25 \color{black}/ \color{blue}0.587&
			21.41 / 0.531&
			15.54 / 0.216&
			\color{red}22.27 \color{black}/  \color{red}0.593\\

			\hline
		\end{tabular}
	\end{center}
	\label{table: near}
	\vspace{-5mm}
\end{table*}

% Experiments
\section{Experiments}
%% Training data
\subsection{Training Data}

We train the proposed DNSR with unpaired real-world HR and LR images. 
%
%These two datasets are independently collected, \ie, there is no correspondence for images from the two different datasets. 
%
Specifically, the HR images are from the DIV2K dataset (with 800 training images) \cite{agustsson2017ntire} and Flickr2K (with 2650 training images) dataset from flickr.com, 
%WQ: is there other LR data?
while we collect the low-quality images from the dataset of Widerface \cite{yang2016wider}, which consists of various LR images of human urban life with unknown degradation and noise. 
%WQ: you need to give the detailed information about training data. numbers.
We select 1600 real-world LR images from Widerface, and randomly crop each real-world LR image with the same size as the generated realistic LR image instead of manually scaling it, which will preserves the original characteristic of real-world LR images.

\subsection{Training Details}

As shown in Figure~\ref{fig 2 framework}, the training process of our algorithm can be divided into three subproblems which are trained iteratively. 
First, we train the degradation model and degradation discriminator with real-world HR images ${x}$ and LR images ${z}$. For computing ${\mathcal{L}_{\text{per}}}$, we scale ${\hat{y}}$ and ${x}$ to the size of 224 $\times$ 224, which is the input size of the first layer of VGG19 network. 
Second, we train the reconstruction model using the generated realistic LR images ${\hat{y}}$. 
Finally, we take the real-world image ${{z}}$ as the input of reconstruction model, and subsequently the generated HR image ${{\tilde z}}$ will be degraded to realistic LR image ${\hat{z}}$ , which is enforced to be similar to the
real-world LR image ${{z}}$.  
For the parameters in Eq.~(\ref{deg loss}) and Eq.~(\ref{rec loss}), we set ${\alpha}$ = 1, ${\beta}$ = 0.5, ${\eta}$ = 1, and ${\gamma}$ = 0.01. 
The minibatch size is set to 16 and HR image size is set to ${240 \times 240}$ pixels. The size of LR images depends on the scale factor, which is set to 2 and 4, respectively.  
The learning rate is initialized as ${10^{-4}}$ and decreased by a factor of 2 every ${2 \times 10^{5}}$ minibatch updates for total ${10^{6}}$ iteration. 
We optimize the total loss function ${{\mathcal{L}_{\text{total}}}}$ with ADAM optimizer \cite{kingma2014adam} by setting ${\beta_1}$ = 0.9 and weight decay to ${10^{-4}}$. 

We implement the proposed method with TensorFlow platform on NVIDIA TITAN X GPUs, and it takes about 2 days to train our model with the scale factor of 2.

%% 4.3 Evaluate on Bicubic Degradation ()
\subsection{Evaluation of Bicubic Degradation}

Although our main goal is to learn a reconstruction model that can deal with real-world image super-resolution, it is difficult to obtain the ground-truth HR images for evaluating the results. 
Therefore, to verify the effectiveness of our method, we first compare our reconstruction model with other CNN-based SISR methods, which are specifically designed for super-resolution based on bicubic degradation. 
The experiments are conducted on five benchmark datasets including Set5 \cite{bevilacqua2012low}, Set14 \cite{zeyde2010single}, Urban100 \cite{huang2015single}, BSD100 \cite{martin2001database}, and DIV2K (with 100 validation images) \cite{agustsson2017ntire}. Each image is down-sampled by bicubic degradation with scale factors of 2 and 4. 
Table~\ref{table: bicu} presents the quantitative results of ours and 5 state-of-the-art methods, including LapSRN\cite{lai2017deep}, DBPN \cite{haris2018deep}, SRMD \cite{Zhang_2018_CVPR}, SRGAN \cite{ledig2017photo}, and ESRGAN \cite{wang2018esrgan}.
As shown, our model can obtain competitive performance to DBPN, which is the winner of NTIRE2018 \cite{Timofte_2018_CVPR_Workshops} on classic bicubic ${\times8}$ track and designed exactly for bicubic degradation. 

\subsection{Evaluation of Nearest-Neighbor Degradation}

\begin{figure*}[htbp]
	\begin{center}
		\includegraphics[width=0.9\linewidth]{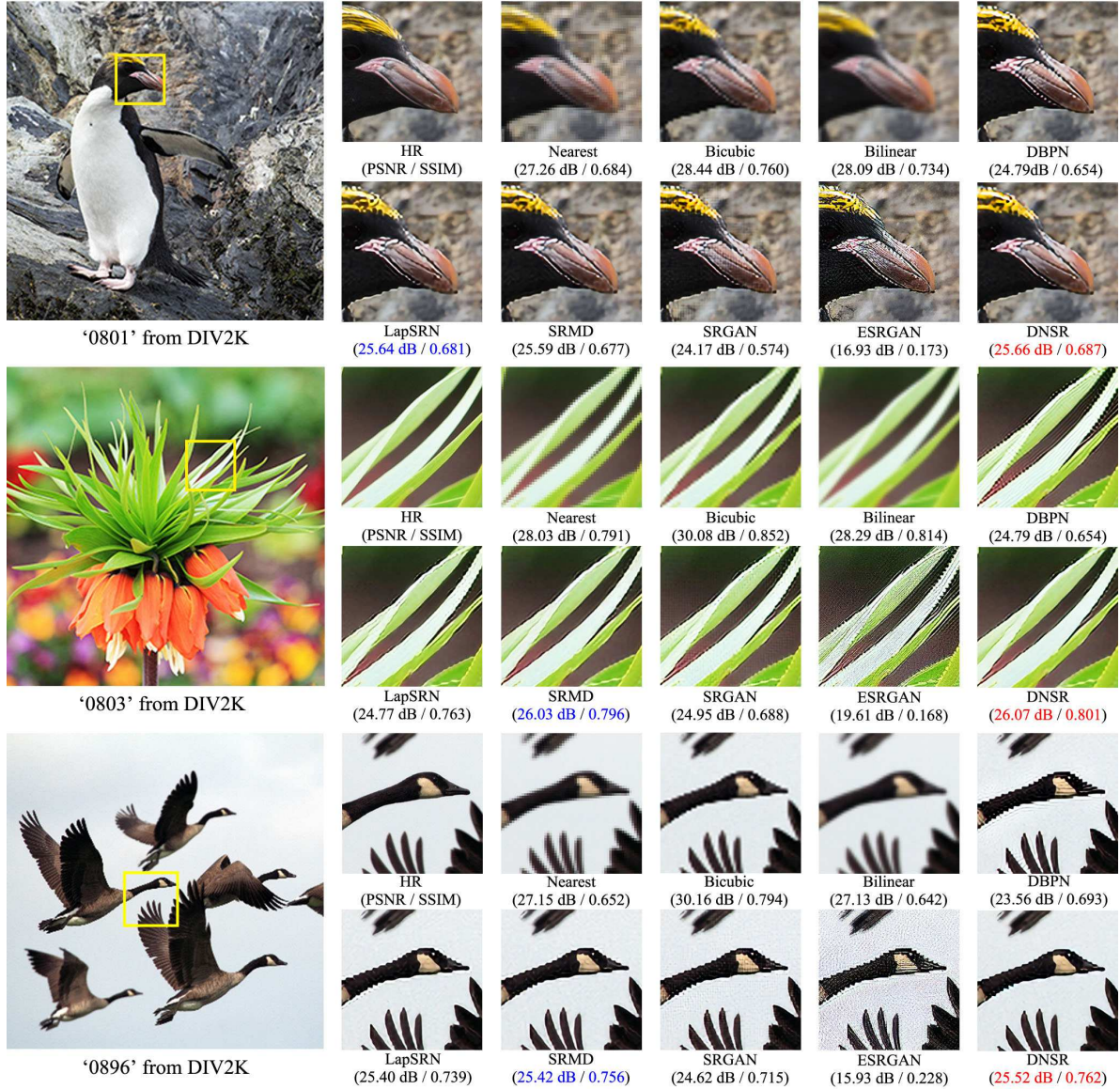}
	\end{center}
	\vspace{-1mm}
	\caption{Qualitative comparison (${\times4}$ SR) between the proposed DNSR model and other state-of-the-arts on nearest-neighbor down-sampled images. The values in red and blue indicate the best and second performances, respectively. It is observed that there are fewer artifacts in the images reconstructed by our model.}
	\label{fig 5 Qualitative comparison}
	\vspace{+2mm}
\end{figure*}

To further evaluate the effectiveness of the proposed method for LR images obtained by different degradation way, we generated LR images by nearest-neighbor degradation with scale factors of 2 and 4, and evaluate our method on the five benchmark datasets.
Table~\ref{table: near} shows the average performance in terms of PSNR and SSIM. 
When compared with LapSRN \cite{lai2017deep}, DBPN \cite{haris2018deep}, SRMD \cite{Zhang_2018_CVPR}, SRGAN \cite{ledig2017photo} and ESRGAN \cite{wang2018esrgan}, our DNSR achieves the best performance on all datasets. 
This 
As shown in Fig.~\ref{fig 5 Qualitative comparison}, although bicubic/bilinear/nearest-neighbor interpolations obtains higher PSNR than the state-of-the-art SISR methods, the results contains obvious blurry edges and textures.
There are obvious artifacts around the edge of objects for the high-resolution images reconstructed by DBPN, LapSRN and SRMD.
Although ESRGAN generates more realistic and natural textures than SRGAN  on the degradation of bicubic, the textures generated tend to be unreal for different types of degradations. 
In contrast, there are fewer artifacts and sharper textures in the super-resolved images generated by our algorithm.

\subsection{Evaluation of Real Images}
%Since we aim to reconstruct the HR images from real-world LR image though there are no ground-truth, 

%%%% figure Qualitative comparison on nearest-neighbor down sampled
\begin{figure*}[t]
	\begin{center}
		\includegraphics[width=0.9 \linewidth]{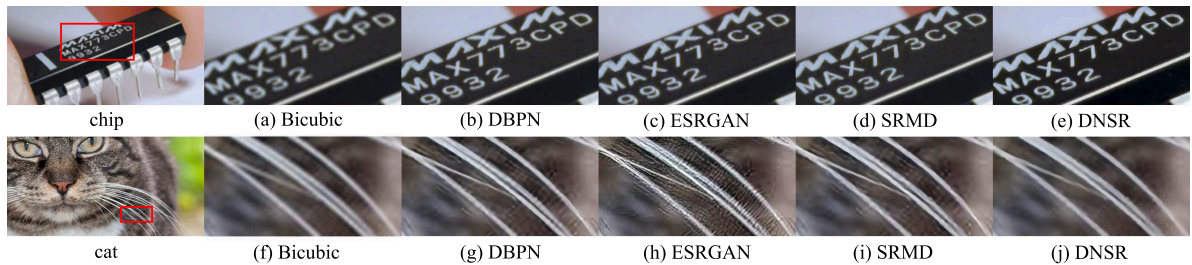}
	\end{center}
	\vspace{-1mm}
	\caption{${\times4}$ SR results on real-world images of \emph{chip} and \emph{cat}. It is observed that there are  sharper edges and fewer artifacts in the images reconstructed by our model.}
	\label{fig 6 real}
	\vspace{-2mm}
\end{figure*}

In this section, we evaluate our reconstruction model on the real-world LR image \emph{chip} (with ${244 \times 200}$ pixels) and \emph{cat} (with ${429 \times 380}$ pixels). Similar to the image shown in Fig.~\ref{fig 1}, both the high-resolution image and the degradation pattern for \emph{chip} and \emph{cat} are unknown, which makes the task rather challenging. As shown in Fig.~\ref{fig 6 real}, we compare our method with Bicubic (as ground-truth), DBPN \cite{haris2018deep}, ESRGAN \cite{wang2018esrgan} and SRMD \cite{Zhang_2018_CVPR}.  For real-world image \emph{chip},  our DNSR recovers sharper edges of characters. The format of real-world LR image \emph{cat} is `jpg', which consists of various artifacts with unknown degradation and noise. As shown in Fig.~\ref{fig 6 real} (g)-(i), the model trained with synthetic images generate more artifacts at the edge of the cat's beard. Benefiting from training with generated realistic LR images, our reconstruction model performs better than the compared ones in restoring sharper edges and less artifacts.

%, DNPB, SRMD and ESRGAN are the methods training with synthetic LR images. 

\section{Analysis and Discussion}
\subsection{Ablation Study}
%%%% figure ablation study result
%%%% figure ablation framework
\begin{figure*}[t]
	\begin{center}
		\includegraphics[width=0.9 \linewidth]{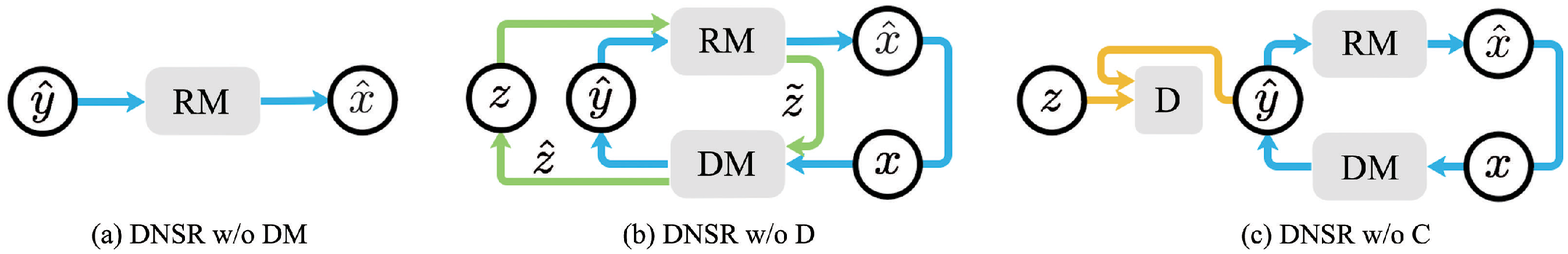}
	\end{center}
	\vspace{-3mm}
	\caption{Three models used in our ablation experiments. RM, DM and D indicate reconstruction model, degradation model and degradation discriminator, respectively. The definitions for the notations used here are the same with those in Fig. 2.}
	\label{fig ablation framework}
	\vspace{-3mm} 
\end{figure*}
\begin{figure*}[t]
	\begin{center}
		\includegraphics[width=0.9 \linewidth]{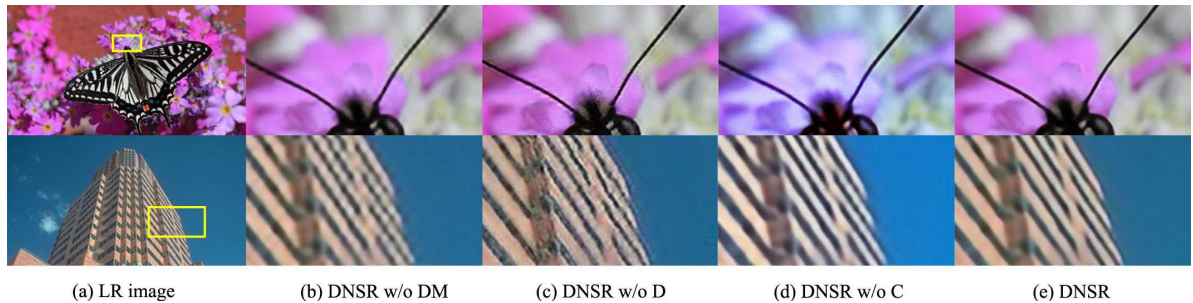}
	\end{center}
	\vspace{-3mm}
	\caption{${\times 4}$ SR results on `0882' (DIV2K) and `img\_095' (BSD100) with different frameworks shown in Fig.~\ref{fig ablation framework}. Our DNSR outperforms the other 3 frameworks with the sharper edge and more realistic color.}
	\label{fig ablation results}
	\vspace{-5mm}
\end{figure*}

%%%% figure noise level
\begin{figure}[]
	\begin{center}
		\includegraphics[width=1 \linewidth]{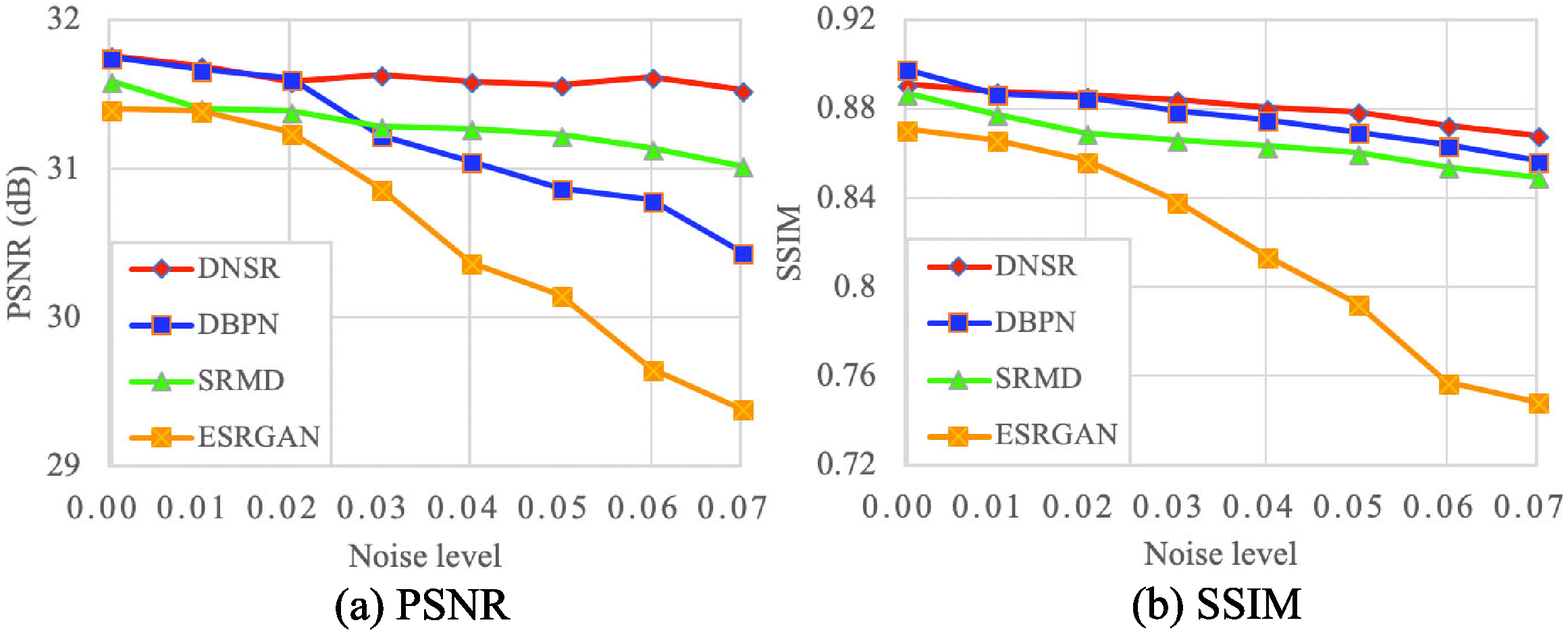}
	\end{center}
	\vspace{-3mm}
	\caption{Quantitative evaluations of several SR methods on the Set5 dataset for ${\times4}$ SR with different noise level.}
	\label{fig noise level}
	\vspace{-5mm}
\end{figure}

%\noindent
%{\bf Framework 1:} 
 
To thoroughly investigate the effectiveness of the proposed method, we conduct ablation experiments by removing specific components for comparison, which induces three different framework as shown in Fig.~\ref{fig ablation framework}. 
%We report the results in Fig.~\ref{fig ablation results}. 
 
 The first framework, \ie,~\textbf{DNSR~w/o~DM}, trains reconstruction model  using bicubic degraded LR images ${\hat y}$, which is corresponding to only minimizing the reconstruction loss ${\mathcal{L}_{\text{rec}}}$ (without ${\mathcal{L}_{\text{cyc}}}$) in Eq.~(\ref{rec loss}). It is observed that the results corresponding \textbf{DNSR~w/o~DM} still have some artifacts at edges comparing with the results generated by our reconstruction model, as shown in Fig.~\ref{fig ablation results}(b) and (e). 
For the second framework, \ie,~\textbf{DNSR~w/o~D}, which removes degradation discriminator model and trains a reconstruction model using realistic LR images ${\hat y}$ generated by the degradation model and the real-world LR images ${z}$, it minimizes the total loss ${\mathcal{L}_{\text{total}}}$ (without ${\mathcal{L}_{\text{adv}}}$) in Eq.~(\ref{total loss}).  The results is obviously unpromising as shown in Fig.~\ref{fig ablation results}(c). Because it is difficult for the degradation model to generate realistic LR images without the degradation discriminator, the reconstruction model performs unpromising. 
For the third framework, \ie,~{\textbf{DNSR~w/o~C}}, which removes the circle ${z}$ ${\rightarrow}$ ${\tilde z}$ ${\rightarrow}$ ${\hat z}$ ${\rightarrow}$ ${z}$ and trains the reconstruction model using ${\hat y}$. We minimize the total loss ${\mathcal{L}_{\text{total}}}$ (without ${\mathcal{L}_{\text{cyc}}}$ both in ${\mathcal{L}_{\text{deg}}}$ and ${\mathcal{L}_{\text{rec}}}$). As shown in Fig.~\ref{fig ablation results}(d), the generated HR images are rather different from the original images in color, which validates the importance of introducing the cycle training strategy to guarantee the consistency across both reconstruction model and degradation model. 

\subsection{Robustness to Noise}
Our proposed DNSR is robust to noisy images. To evaluate the robustness of DNSR, we down-sample the test images and randomly add Gaussian noise with noise level $\sigma$ from 1\% to 7\% on the Set5 dataset. Fig.~\ref{fig noise level} shows quantitative results of some state-of-the-art methods on the test dataset with scale factor of 4. 
As the noise level increases, the performances of these methods decrease with different extents. It is observed that ESRGAN is rather sensitive to noise. The possible reason is that ESRGAN aims to generate realistic images  with emphasizing more on textures and less on noise, which makes the noise is strengthened as textures. Due to the unsupervised degradation network for generating realistic LR images, our method performs much better with increased noise level.

\subsection{Running Time}

\begin{table}[t]
\caption{Average running time (in second) on the dataset DIV2K.}
\vspace{-5mm}
%\scriptsize
%\tiny
\scriptsize
\begin{center}
\begin{tabular}{|c|c|c|c|c|c|c|}
\hline
\multirow{1}*{Platform} & 
\multicolumn{2}{|c|}{MATLAB} &	
\multicolumn{2}{|c|}{PyTorch}&	
\multicolumn{2}{|c|}{TensorFlow} \\
\hline
Method & LapSRN & SRMD & DBPN & ESRGAN & SRGAN & DNSR \\
\hline
%PSNR & 28.75 & 28.72 & 28.94 & 28.68 & 28.09 & 28.87 \\
%\hline
\multirow{1}*{Time} & 
8.433&	0.697&	21.64&	3.31&	0.657&	\bf0.645 \\
\hline
\end{tabular}
\end{center}
\label{table: run time}
\vspace{-8mm}
\end{table}

Table~\ref{table: run time} shows the time cost for different methods on the benchmark dataset DIV2K. We test all the methods on NVIDIA TITAN X GPUs with the scale factor of 4. Though there are some differences in terms of computing ability between the implementation platforms (\ie~Matlab, PyTorch, and TensorFlow), 
it can be observed that our method achieves rather promising performance in terms of efficiency.

\section{Conclusion}

%In this paper, we propose an unsupervised degradation network for single image super-resolution. The proposed method first learns to reconstruct realistic low-resolution images by using the degradation model, then super-resolves images via the reconstruction model based on generated realistic low-resolution images, which makes the proposed DNSR is able to recover images degraded by multiple degradation types. 
%of degraded images demonstrate that our model has a strong ability in handling various degradation patterns comparing with other supervised super-resolution methods. 
%
%WQ: real, real, real...............
%Moreover, our model outperforms the state-of-the-art algorithms on reconstructing real-world images with more realistic textures and colors. In the future we will design more powerful network structure and employ more data to further promote the performance and validates the advantages.

In this paper, we propose an unsupervised degradation network for single image super-resolution, which does not need paired manually generated low-resolution images as the ground-truth. The proposed method jointly learns to generated realistic low-resolution images with a degradation module and super-resolve high-resolution images with the reconstruction network based on generated realistic low-resolution images, which endows with the proposed DNSR the ability of super-resolving real-world low-resolution images. With extensive experiments, our model outperforms the state-of-the-art algorithms for reconstructing real-world images. In the future, we will design more powerful network and employ more diverse data to further investigate the potential of performance promotion.

%-------------------------------------------------------------------------

\clearpage
{\small
\bibliographystyle{ieee}
\bibliography{egbib}
}

\end{document}